\documentclass{article}

\usepackage{PRIMEarxiv}

\usepackage[utf8]{inputenc} % allow utf-8 input
\usepackage[T1]{fontenc}    % use 8-bit T1 fonts
\usepackage{float}
\usepackage{hyperref}       % hyperlinks
\usepackage{url}            % simple URL typesetting
\usepackage{booktabs}       % professional-quality tables
\usepackage{amsfonts}       % blackboard math symbols
\usepackage{nicefrac}       % compact symbols for 1/2, etc.
\usepackage{microtype}      % microtypography
\usepackage{lipsum}
\usepackage{fancyhdr}       % header
\usepackage{graphicx} 
\usepackage{tikz}
\usetikzlibrary{positioning, arrows.meta}
\usepackage{amsmath} % graphics\
\usepackage{cleveref}
\graphicspath{{media/}}     % organize your images and other figures under media/ folder

\title{AdamZ: An Enhanced Optimisation Method for Neural Network Training
}

\author{
  Ilia Zaznov, Atta Badii \\
  Department of Computer Science \\
  University of Reading \\
  Reading, UK \\
  \texttt{\{i.zaznov@pgr.reading.ac.uk, atta.badii@reading.ac.uk\}} \\
  \And
  Julian Kunkel \\
  Department of Computer Science/GWDG \\
  University of Göttingen \\
  Goettingen, Germany \\
  \texttt{julian.kunkel@gwdg.de} \\
  \And
  Alfonso Dufour \\
  ICMA Centre, Henley Business School \\
  University of Reading \\
  Reading, UK \\
  \texttt{a.dufour@icmacentre.ac.uk} \\
}

\begin{document}
\maketitle

\begin{abstract}
AdamZ is an advanced variant of the Adam optimiser, developed to enhance convergence efficiency in neural network training. This optimiser dynamically adjusts the learning rate by incorporating mechanisms to address overshooting and stagnation, that are common challenges in optimisation. Specifically, AdamZ reduces the learning rate when overshooting is detected and increases it during periods of stagnation, utilising hyperparameters such as overshoot and stagnation factors, thresholds, and patience levels to guide these adjustments. While AdamZ may lead to slightly longer training times compared to some other optimisers, it consistently excels in minimising the loss function, making it particularly advantageous for applications where precision is critical. Benchmarking results demonstrate the effectiveness of AdamZ in maintaining optimal learning rates, leading to improved model performance across diverse tasks.

\end{abstract}

% keywords can be removed

\section{Introduction}
In recent years, the field of machine learning has witnessed significant advancements, particularly in the development of optimisation algorithms that enhance the efficiency and effectiveness of training deep neural networks. Among these algorithms, the Adam optimiser has gained widespread popularity due to its adaptive learning rate capabilities, which enable more efficient convergence compared to traditional methods such as stochastic gradient descent. However, despite its advantages, Adam is not without its limitations, particularly when it comes to handling issues such as overshooting and stagnation during the training process.

To address these challenges, we introduce AdamZ as an advanced variant of the Adam optimiser. AdamZ is specifically designed to dynamically adjust the learning rate responsive to the characteristics of the loss function, thereby improving both convergence stability and model accuracy. This novel optimiser integrates mechanisms to detect and mitigate overshooting, at the point where the optimiser has stepped too far into the parameter space, and stagnation at points, where progress has started to stall despite ongoing training. By introducing hyperparameters such as overshoot and stagnation factors, thresholds, and patience levels, AdamZ provides a more responsive approach to learning rate adaptation than obtained through Adam.

The development of AdamZ is motivated by the need for more robust optimisation techniques that can adaptively respond to the dynamic nature of neural network training landscapes. Traditional fixed learning rate strategies often struggle with the non-convex and irregular loss surfaces characteristic of deep learning problems, leading to suboptimal performance. By contrast, the AdamZadaptive strategy enables more precise navigation of these complex landscapes, reducing the likelihood of getting trapped in local minima or experiencing erratic convergence behaviour.

In this paper, we present a comprehensive analysis of the AdamZ optimiser, detailing its theoretical underpinnings and practical implementation. We also provide empirical evidence demonstrating its superior performance in terms of loss minimisation and accuracy of model predictions across a range of benchmark datasets, albeit with slightly longer training times compared to some other optimisers. This trade-off highlights the potential of AdamZ as a valuable tool for applications where accuracy is paramount.

The remainder of this paper is structured as follows: Section 2 reviews related work in optimisation algorithms for deep learning. Section 3 details the implementation of AdamZ, including its key features and hyperparameters. Section 4 presents experimental results comparing AdamZ to other leading optimisers. Finally, Section 5 discusses the implications of our findings and suggests directions for future research.

\section{Related Work}
The development of optimisation algorithms has been pivotal in advancing the field of machine learning, particularly in training deep neural networks. This section reviews several key optimisers that have influenced the design and functionality of AdamZ. \Cref{tab:optimisers} summarises these optimisers, highlighting their distinctive features and contributions to the field.

\begin{table}[htbp]
    \centering
    \begin{tabular}{|l|l|c|p{7cm}|p{2.5cm}|p{2.5cm}|}
        \hline
        \textbf{Optimiser} & \textbf{Reference} & \textbf{Year} & \textbf{Description} & \textbf{Hyperparameters} & \textbf{Application} \\ \hline
        SGD & \cite{robbins1951stochastic} & 1951 & Stochastic Gradient Descent (SGD) updates parameters iteratively by moving in the direction of the negative gradient of the loss function. The update rule is given by: 
        \[
        \theta_{t+1} = \theta_t - \eta \nabla_\theta J(\theta)
        \]
        where $\eta$ is the learning rate. & 1 (learning rate) & Requires careful tuning of learning rate. \\ \hline
        ASGD & \cite{polyak1992acceleration} & 1992 & Averaged Stochastic Gradient Descent (ASGD) improves convergence by averaging the sequence of iterates. The update rule is:
        \[
        \bar{\theta}_t = \frac{1}{t} \sum_{i=1}^{t} \theta_i
        \]
        & 1 (learning rate) & Effective in reducing variance in updates. \\ \hline
        Adagrad & \cite{duchi2011adaptive} & 2011 & Adagrad adapts the learning rate for each parameter based on historical gradients. The update rule is:
        \[
        \theta_{t+1} = \theta_t - \frac{\eta}{\sqrt{G_t + \epsilon}} \nabla_\theta J(\theta)
        \]
        where $G_t$ is the sum of squares of past gradients and $\epsilon$ is a small constant. & 1 (learning rate) & Accumulates squared gradients, which can lead to overly small learning rates. \\ \hline
        RMSprop & \cite{tieleman2012lecture} & 2012 & RMSprop modifies Adagrad by introducing a decay factor to control the accumulation of past gradients:
        \[
        E[g^2]_t = \gamma E[g^2]_{t-1} + (1-\gamma)g_t^2
        \]
        \[
        \theta_{t+1} = \theta_t - \frac{\eta}{\sqrt{E[g^2]_t + \epsilon}} \nabla_\theta J(\theta)
        \]
        & 2 (learning rate, decay rate) & Commonly used in recurrent neural networks. \\ \hline
        Adam & \cite{kingma2014adam} & 2014 & Adam combines the advantages of Adagrad and RMSprop, using moving averages of the gradient and its square:
        \[
        m_t = \beta_1 m_{t-1} + (1-\beta_1)g_t
        \]
        \[
        v_t = \beta_2 v_{t-1} + (1-\beta_2)g_t^2
        \]
        \[
        \hat{m}_t = \frac{m_t}{1-\beta_1^t}, \quad \hat{v}_t = \frac{v_t}{1-\beta_2^t}
        \]
        \[
        \theta_{t+1} = \theta_t - \frac{\eta}{\sqrt{\hat{v}_t} + \epsilon} \hat{m}_t
        \]
        & 3 (learning rate, beta1, beta2) & Widely used due to its robustness and efficiency. \\ \hline
        AdaMax & \cite{kingma2014adam} & 2014 & AdaMax is a variant of Adam using the infinity norm:
        \[
        u_t = \max(\beta_2 u_{t-1}, |g_t|)
        \]
        \[
        \theta_{t+1} = \theta_t - \frac{\eta}{u_t} \hat{m}_t
        \]
        & 3 (learning rate, beta1, beta2) & Particularly useful for models with large parameter spaces. \\ \hline
        NAdam & \cite{dozat2016incorporating} & 2016 & NAdam incorporates Nesterov momentum into Adam:
        \[
        \theta_{t+1} = \theta_t - \frac{\eta}{\sqrt{\hat{v}_t} + \epsilon} (\beta_1 \hat{m}_{t-1} + (1-\beta_1)g_t)
        \]
        & 3 (learning rate, beta1, beta2) & Suitable for non-convex optimisation problems. \\ \hline
        AdamW & \cite{loshchilov2017decoupled} & 2017 & AdamW decouples weight decay from the gradient-based update:
        \[
        \theta_{t+1} = \theta_t - \frac{\eta}{\sqrt{\hat{v}_t} + \epsilon} \hat{m}_t - \eta \lambda \theta_t
        \]
        & 4 (learning rate, beta1, beta2, weight decay) & Suitable for training large models with regularization. \\ \hline
    \end{tabular}
    \caption{Summary of Key Optimisers}
    \label{tab:optimisers}
\end{table}

Stochastic Gradient Descent (SGD) \cite{robbins1951stochastic} is a foundational optimisation algorithm introduced in 1951. It updates parameters iteratively by moving in the direction of the negative gradient of the loss function. The update rule is given by:
\[
\theta_{t+1} = \theta_t - \eta \nabla_\theta J(\theta)
\]
where \(\eta\) is the learning rate. This method is commonly used in training machine learning models, especially in deep learning. It is simple to implement and computationally efficient for large datasets. However, it requires careful tuning of the learning rate and can get stuck in local minima.

Averaged Stochastic Gradient Descent (ASGD) \cite{polyak1992acceleration}, introduced in 1992, enhances convergence by averaging the sequence of iterates. The update rule is:
\[
\bar{\theta}_t = \frac{1}{t} \sum_{i=1}^{t} \theta_i
\]
This approach is effective in reducing variance in updates and is used in scenarios where reducing the variance of SGD updates is crucial, such as in online learning. However, the averaging process can slow down convergence in the initial stages.

Adagrad (2011) \cite{duchi2011adaptive} adapted the learning rate for each parameter based on historical gradients. The update rule is:
\[
\theta_{t+1} = \theta_t - \frac{\eta}{\sqrt{G_t + \epsilon}} \nabla_\theta J(\theta)
\]
where \(G_t\) is the sum of squares of past gradients and \(\epsilon\) is a small constant. It is suitable for sparse data and problems with sparse gradients, such as natural language processing tasks. However, it accumulates squared gradients, which can result in a continually decreasing learning rate.

RMSprop (2012) \cite{tieleman2012lecture} modified Adagrad by introducing a decay factor to control the accumulation of past gradients:
\[
E[g^2]_t = \gamma E[g^2]_{t-1} + (1-\gamma)g_t^2
\]
\[
\theta_{t+1} = \theta_t - \frac{\eta}{\sqrt{E[g^2]_t + \epsilon}} \nabla_\theta J(\theta)
\]
This is popular in training recurrent neural networks and models with non-stationary objectives. It requires tuning of the decay factor and learning rate.

Adam (2014) \cite{kingma2014adam} combined the advantages of Adagrad and RMSprop by using moving averages of the gradient and its square:
\[
m_t = \beta_1 m_{t-1} + (1-\beta_1)g_t
\]
\[
v_t = \beta_2 v_{t-1} + (1-\beta_2)g_t^2
\]
\[
\hat{m}_t = \frac{m_t}{1-\beta_1^t}, \quad \hat{v}_t = \frac{v_t}{1-\beta_2^t}
\]
\[
\theta_{t+1} = \theta_t - \frac{\eta}{\sqrt{\hat{v}_t} + \epsilon} \hat{m}_t
\]
This optimiser is widely used across various deep learning applications due to its robust performance. However, it can sometimes lead to non-convergent behaviour or overfitting if not properly tuned.

AdaMax (2014) \cite{kingma2014adam} is a variant of Adam using the infinity norm:
\[
u_t = \max(\beta_2 u_{t-1}, |g_t|)
\]
\[
\theta_{t+1} = \theta_t - \frac{\eta}{u_t} \hat{m}_t
\]
It is particularly useful for models with large parameter spaces.

NAdam (2016) \cite{dozat2016incorporating} incorporated Nesterov momentum into Adam:
\[
\theta_{t+1} = \theta_t - \frac{\eta}{\sqrt{\hat{v}_t} + \epsilon} (\beta_1 \hat{m}_{t-1} + (1-\beta_1)g_t)
\]
This optimiser is effective in non-convex optimisation problems.

AdamW \cite{loshchilov2017decoupled}, introduced in 2017, decouples weight decay from the gradient-based update:
\[
\theta_{t+1} = \theta_t - \frac{\eta}{\sqrt{\hat{v}_t} + \epsilon} \hat{m}_t - \eta \lambda \theta_t
\]
It is particularly useful in training large models with regularization, such as in transformer architectures.

The optimization methods mentioned are not the only ones available, but they are the most widely used and well-established. Other prominent contributions in this field were: \cite{loshchilov2016sgdr}, \cite{reddi2019convergence}, \cite{luo2019adaptive}, \cite{ioannou2023adalip}, \cite{zhang2023adagl}, and \cite{zuo2023nala}

Despite the extensive development of various optimisation algorithms, each optimiser has its own set of challenges and limitations. Techniques such as SGD, Adagrad, RMSprop, and Adam have significantly advanced the training of machine learning models, yet they often struggle with issues such as convergence speed, sensitivity to hyperparameter tuning, and susceptibility to getting stuck in local minima. 
These limitations highlight the necessity for continuous innovation in optimisation methods. The proposed AdamZ optimiser aims to address these issues by enhancing convergence rates and improving the ability to reach global minima. By incorporating novel mechanisms to dynamically adjust learning rates and momentum, AdamZ seeks to provide a more robust and efficient optimisation strategy, paving the way for improved performance in complex, high-dimensional learning tasks. The next section provides a comprehensive overview of the implementation of AdamZ, highlighting its key features and the specific hyperparameters that drive its performance.

\section{Implementation of AdamZ Optimiser}

The AdamZ optimiser is an advanced variant of the traditional Adam optimiser, designed to provide more adaptive learning rate adjustments during the training process. This optimiser addresses two common issues faced in optimisation: overshooting and stagnation. Overshooting occurs when the learning rate is too high, causing the optimiser to miss the optimal point, while stagnation happens when the learning rate is too low, resulting in slow convergence or getting stuck in local minima.

The motivation behind AdamZ is to enhance the flexibility and robustness of the learning process by dynamically adjusting the learning rate based on the behaviour of the loss function. Traditional Adam, while effective, can be deficient in responsively adapting the learning rate given with dynamically changing landscapes of the loss function, leading to inefficient convergence. AdamZ introduces mechanisms to detect and mitigate these issues by adjusting the learning rate in response to overshooting and stagnation, thereby improving the optimiseradaptability and efficiency.

Thus, AdamZ incorporates additional hyperparameters that enable it to respond to the training dynamics:

\begin{itemize}
\item \textbf{Overshoot Factor} ($\gamma_{\text{over}}$): Reduces the learning rate once overshooting has been detected, preventing the optimiser from overshooting the minimum.
\item \textbf{Stagnation Factor} ($\gamma_{\text{stag}}$): Increases the learning rate once loss has started to plateau thus indicating the onset of stagnation is detected, helping the optimiser to escape local minima.
\item \textbf{Stagnation Threshold} ($\sigma_{\text{stag}}$): Determines the sensitivity of the stagnation detection based on the standard deviation of the loss.
\item \textbf{Patience} ($p$): Number of steps to wait before adjusting the learning rate, allowing for a stable assessment of the loss trend.
\item \textbf{Stagnation Period} ($s$): Number of steps over which stagnation is assessed.
\item \textbf{Learning Rate Bounds} ($\alpha_{\text{min}}$, $\alpha_{\text{max}}$): Ensures that the learning rate remains within this specified range to prevent extreme adjustments.
\end{itemize}

The above six parameters essentially enable measured i) agility control, and ii) over-reactivity control of the Adamz optimiser in attempting its dynamic responsive adaptation of the learning rate.

The performance of AdamZ is highly dependent on the careful tuning of its hyperparameters. Each hyperparameter plays a critical role in determining how the optimiser reacts to changes in the loss function. Fine-tuning these parameters through techniques such as grid search can significantly enhance the optimiser's performance by adapting it to the specific characteristics of the problem at hand.

The mathematical formulation and algorithmic steps involved in implementing the AdamZ optimiser are set out below:

\begin{enumerate}
\item \textbf{Initialise Parameters:} Initialise the parameters $\theta_0$, and set initial values for the first moment vector $m_0 = 0$, second moment vector $v_0 = 0$, and step $t = 0$.

\item \textbf{Hyperparameters:} Define the hyperparameters (note that these are default values inferred from our validation experiments, they must be task/domain-specifically fine-tuned for best performance):
\begin{itemize}
    \item Learning rate $\alpha = 0.01$
    \item Exponential decay rates for the moment estimates $\beta_1 = 0.9$, $\beta_2 = 0.999$
    \item Stability constant $\epsilon = 10^{-8}$
    \item Overshoot factor $\gamma_{\text{over}} = 0.5$
    \item Stagnation factor $\gamma_{\text{stag}} = 1.2$
    \item Stagnation threshold $\sigma_{\text{stag}} = 0.2$
    \item Patience period $p = 100$
    \item Stagnation period $s = 10$
    \item Maximum gradient norm $N_{\text{max}} = 1.0$
    \item Minimum and maximum learning rates $\alpha_{\text{min}} = 10^{-7}$, $\alpha_{\text{max}} = 1$
\end{itemize}

\item \textbf{Update Rule:} For each iteration $t$, perform the following steps:
\begin{enumerate}
    \item Compute gradients $g_t = \nabla_{\theta} f_t(\theta_{t-1})$.
    \item Update biased first moment estimate:
    \[
    m_t = \beta_1 \cdot m_{t-1} + (1 - \beta_1) \cdot g_t
    \]
    \item Update biased second raw moment estimate:
    \[
    v_t = \beta_2 \cdot v_{t-1} + (1 - \beta_2) \cdot g_t^2
    \]
    \item Compute bias-corrected first moment estimate:
    \[
    \hat{m}_t = \frac{m_t}{1 - \beta_1^t}
    \]
    \item Compute bias-corrected second raw moment estimate:
    \[
    \hat{v}_t = \frac{v_t}{1 - \beta_2^t}
    \]
    \item Update parameters:
    \[
    \theta_t = \theta_{t-1} - \alpha \cdot \frac{\hat{m}_t}{\sqrt{\hat{v}_t} + \epsilon}
    \]
\end{enumerate}

\item \textbf{Adjust Learning Rate:} Evaluate the current loss $L_t$ and adjust the learning rate:
\begin{itemize}
    \item If overshooting is detected, i.e., $L_t \geq \max(\{L_{t-p}, \ldots, L_t\})$, then:
    \[
    \alpha \leftarrow \alpha \cdot \gamma_{\text{over}}
    \]
    \item If stagnation is detected, i.e., $\text{std}(\{L_{t-s}, \ldots, L_t\}) < \sigma_{\text{stag}} \cdot \text{std}(\{L_{t-p}, \ldots, L_t\})$, then:
    \[
    \alpha \leftarrow \alpha \cdot \gamma_{\text{stag}}
    \]
\end{itemize}
Ensure $\alpha$ is bounded: $\alpha = \max(\alpha_{\text{min}}, \min(\alpha, \alpha_{\text{max}}))$.

\item \textbf{Gradient Clipping:} Clip the gradients to the maximum norm $N_{\text{max}}$.
\end{enumerate}

\begin{figure}[h]
    \centering
    \includegraphics[width=0.7\textwidth]{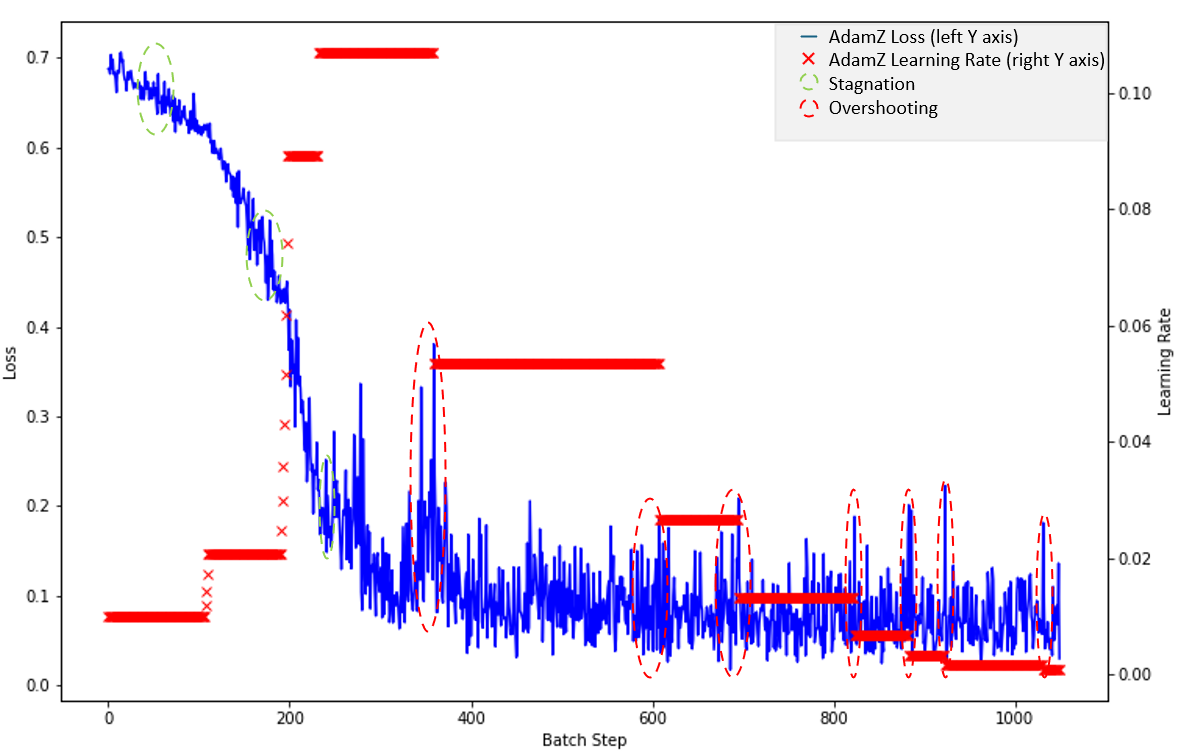}
    \caption{AdamZ mechanism of detecting and responding to overshooting and stagnation}
    \label{fig:stagnation_overshooting}
\end{figure}

The code implementation in Python for the AdamZ optimiser is available at the following GitHub repository:\url{https://github.com/izaznov/AdamZ.git}

\Cref{fig:stagnation_overshooting} illustrates the mechanism of the AdamZ identification of the stagnation and overshooting patterns and respective learning rate adjustments. Initially, periods of stagnation, marked by losses fluctuating around a constant value, trigger three rounds of learning rate increase. Subsequently, spikes in losses indicate overshooting, which in turn trigger seven rounds of learning rate reduction.

The introduction of dynamic learning rate adjustments in AdamZ enables it to handle a wider range of optimisation challenges compared to traditional methods. By incorporating mechanisms to detect and respond to overshooting and stagnation, with measured agility to avoid over-reactivity as restrained by the aforementioned control thresholds, AdamZ enhances the convergence speed and stability of the training process. This makes it particularly valuable in complex neural network training scenarios where traditional optimisers may falter.

In conclusion, AdamZ represents a significant advancement in optimisation techniques, offering a more measured and responsive approach to learning rate adjustment. Its effectiveness, however, is contingent upon the appropriate tuning of its hyperparameters, underscoring the importance of experimental validation and parameter optimisation.

\section{Experimental Results}

\subsection{Objectives}
The primary objective of these experiments was to evaluate the performance of the proposed optimiser, AdamZ, in comparison with other popular optimisation algorithms. We aimed to assess its effectiveness across two different datasets, including a synthetic dataset generated using \texttt{make\_circles} (from sklearn.datasets) and the widely-used MNIST dataset (from torchvision). The implementation of these experiments in Python is available in GitHub repository:\url{https://github.com/izaznov/AdamZ.git}, with the first experiment in the file \texttt{Circle\_adamz\_whitepaper.py} and the second one in: \texttt{Mnist\_adamz\_whitepaper.py}.

These experiments provided insights into the optimiser's performance, including loss, accuracy of prediction, training duration, and overall applicability in neural network training.

\subsection{Experiment 1: Synthetic Dataset Using \texttt{make\_circles} with a shallow neural network} 

This experiment utilised the \texttt{make\_circles} function from sklearn.datasets to generate a synthetic dataset for binary classification tasks. The controlled nature of this dataset enabled a clear assessment of the optimiser's performance. 
The \texttt{make\_circles} function in the Scikit-learn library was used to generate a synthetic dataset that is particularly useful for binary classification problems. This dataset consisted of two interleaving circles, which made it a challenging test case for algorithms that rely on linear separability. The dataset was generated in a two-dimensional space and was often used to demonstrate the capabilities of non-linear classifiers such as support vector machines with a radial basis function kernel or neural networks.

The \texttt{make\_circles} function enabled several parameters to be specified, such as:
\begin{itemize}
    \item \texttt{n\_samples}: This parameter defined the total number of samples to be generated.
    \item \texttt{noise}: This parameter specified the standard deviation of Gaussian noise to be added to the data, which can help simulate real-world conditions.
    \item \texttt{factor}: This was the scale factor between the inner and outer circle, determining the relative size of the circles.
\end{itemize}

\begin{figure}[h]
    \centering
    \includegraphics[width=0.7\textwidth]{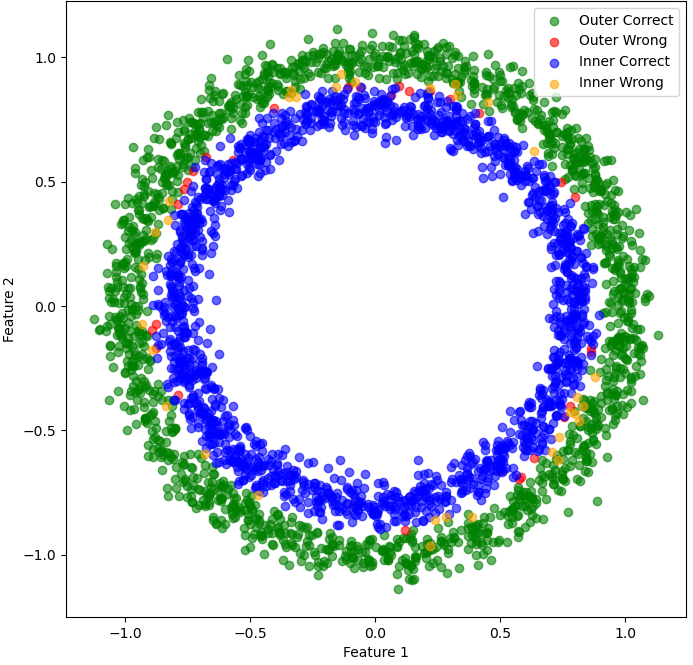}
    \caption{Visualisation of the make\_circles dataset }
    \label{fig:make_circles}
\end{figure}

Using this dataset, one can effectively test and visualise the performance of classification algorithms that are designed to handle non-linear decision boundaries. \Cref{fig:make_circles} illustrates a typical visualisation of the \texttt{make\_circles} dataset, showing the inner and outer circles and illustrating the ability of the model to correctly classify whether the particular point belonged to the inner or outer circle.

\begin{figure}[h]
    \centering
    \begin{tikzpicture}[
        node distance=2.5cm,
        every node/.style={draw, minimum width=2cm, minimum height=1cm, font=\small},
        input/.style={fill=green!20, rectangle},
        hidden/.style={fill=blue!20, rectangle},
        output/.style={fill=red!20, rectangle},
        arrow/.style={-{Stealth}}
    ]

    % Nodes
    \node[input] (input) {\begin{tabular}{c} Input \\ (2) \end{tabular}};
    \node[right=of input, hidden] (hidden) {\begin{tabular}{c} Hidden \\ (10) ReLU \end{tabular}};
    \node[right=of hidden, output] (output) {\begin{tabular}{c} Output \\ (1) Sigmoid \end{tabular}};

    % Layers
    \node[above=0cm of hidden, align=center, draw=none] (layer1) {Layer 1: $\text{Linear}(2, 10)$};
    \node[above=0cm of output, align=center, draw=none] (layer2) {Layer 2: $\text{Linear}(10, 1)$};

    % Connections
    \draw[arrow] (input) -- (hidden);
    \draw[arrow] (hidden) -- (output);

    \end{tikzpicture}
    \caption{A neural network architecture applied to the make\_circles dataset.}
    \label{fig:neural_network}
\end{figure}
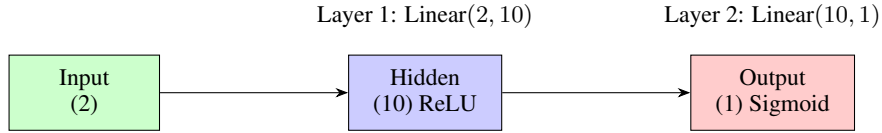

A shallow neural network architecture was employed, consisting of several fully connected layers. This setup is ideal for quick experimentation and testing on smaller datasets. \Cref{fig:neural_network} shows the architecture of the neural network deployed. The hyperparameters were carefully chosen to ensure a fair comparison across different optimisers.

The performance of AdamZ was compared against well-established optimisers such as Adam, SGD, and RMSprop, providing a benchmark for evaluating improvements in terms of model classification accuracy, training time, and loss minimisation for 10 epochs (~100 steps in each epoch) of training with 100 simulations to account for randomness in the parameter initialisation. The experiment was conducted on a compute cluster with 4x A100 GPUs. 

As illustrated in \Cref{fig:make_circles}, and summarised in \Cref{tab:synthetic_results}, AdamZ demonstrated the highest classification accuracy (\%), but slightly longer training time (measured in seconds) compared to the other optimisers.

\begin{table}[H]
    \centering
    \begin{tabular}{|l|ccc|ccc|}
        \hline
        \textbf{Optimiser} & \multicolumn{3}{c|}{\textbf{Accuracy}} & \multicolumn{3}{c|}{\textbf{Training Duration}} \\
        \cline{2-7}
        & \textbf{Q1} & \textbf{Median} & \textbf{Q4} & \textbf{Q1} & \textbf{Median} & \textbf{Q4} \\
        \hline
        Adam & 97.58 & 97.76 & 97.85 & 1.73 & 1.74 & 1.75 \\
        AdamW & 97.58 & 97.73 & 97.86 & 1.76 & 1.77 & 1.78 \\
        SGD & 54.21 & 56.32 & 58.84 & 1.47 & 1.48 & 1.48 \\
        RMSprop & 97.45 & 97.67 & 97.82 & 1.65 & 1.66 & 1.66 \\
        Adagrad & 59.51 & 61.80 & 64.50 & 1.66 & 1.67 & 1.67 \\
        Adamax & 95.39 & 96.33 & 96.83 & 1.69 & 1.70 & 1.71 \\
        ASGD & 53.19 & 56.14 & 58.92 & 2.48 & 2.50 & 2.51 \\
        NAdam & 97.51 & 97.70 & 97.88 & 1.84 & 1.84 & 1.85 \\
        \textbf{AdamZ} & 97.67 & 97.83 & 98.00 & 3.01 & 3.02 & 3.03 \\
        \hline
    \end{tabular}
    \caption{Performance comparison of optimisers on the make\_circles dataset.}
    \label{tab:synthetic_results}
\end{table}

\begin{figure}[h]
    \centering
    \includegraphics[width=0.8\textwidth]{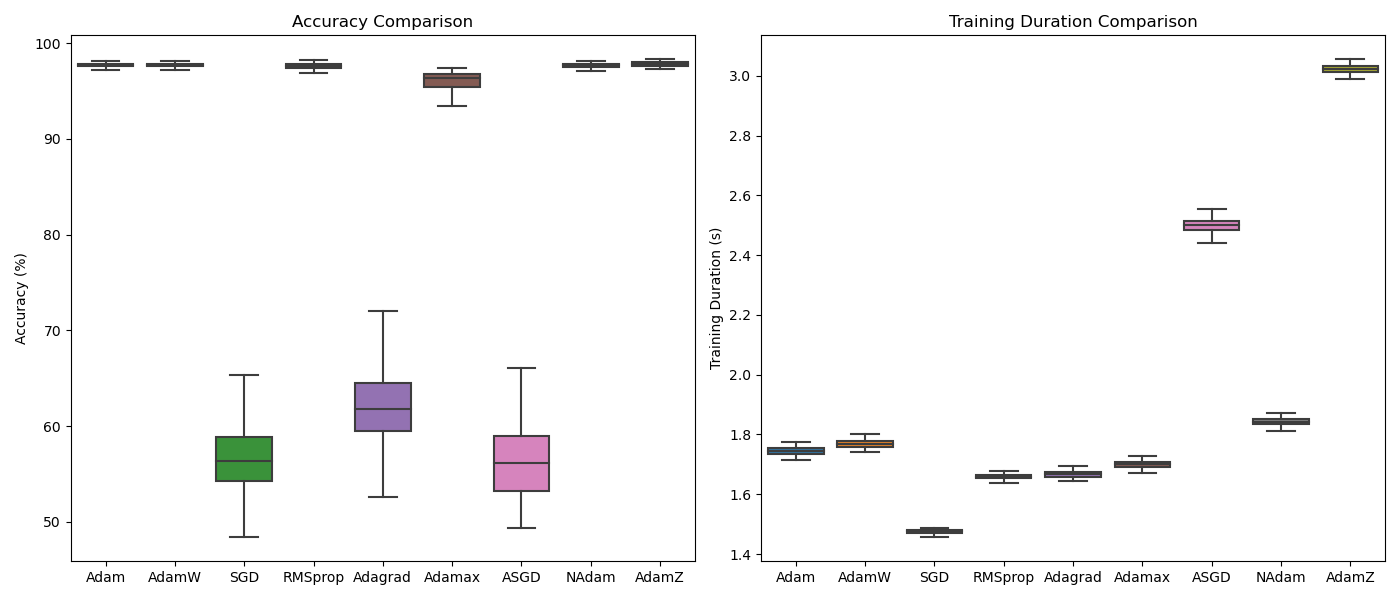}
    \caption{Accuracy and training duration on the make\_circles dataset.}
    \label{fig:comparison_charts_synthetic}
\end{figure}

\begin{figure}[h]
    \centering
    \includegraphics[width=0.8\textwidth]{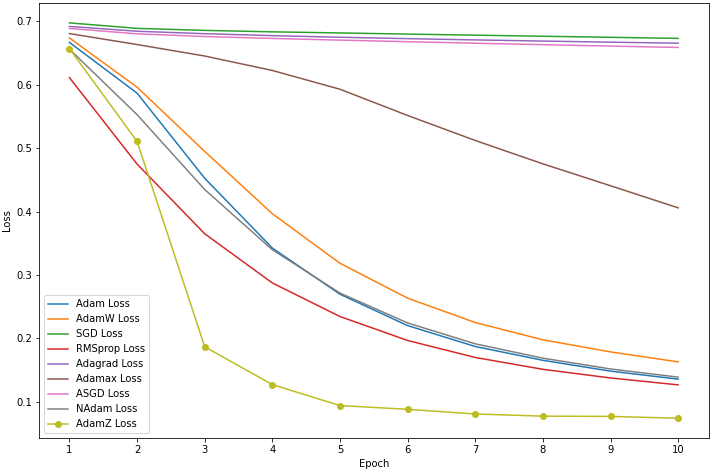}
    \caption{Optimzers' loss evolution for the make\_circles dataset.}
    \label{fig:circle_loss}
\end{figure}

Considering the training loss evolution of the optimisers in \Cref{fig:circle_loss}, it can be seen that AdamZ is also minimising loss better than other optimisers.

\subsection{Experiment 2: MNIST Dataset with a deep neural network}

The MNIST dataset is a cornerstone in the field of machine learning and computer vision, widely used for training various image processing systems. It was introduced by Yann LeCun, Corinna Cortes, and Christopher J.C. Burges \cite{lecun1998gradient}. MNIST stands for Modified National Institute of Standards and Technology database, and it is a large collection of handwritten digits that is commonly used to train image processing systems.

The dataset consists of 70,000 images of handwritten digits from 0 to 9. These images are grayscale and normalised to fit in a 28x28 pixel bounding box, preserving the aspect ratio of the original digit. This normalisation process ensures that the dataset is consistent and easy to use for various machine-learning models.

Key features of the MNIST dataset include:
\begin{itemize}
    \item \textbf{Training Set}: 60,000 images used for training models.
    \item \textbf{Test Set}: 10,000 images used for evaluating model performance.
    \item \textbf{Balanced Classes}: Each digit class is represented equally, providing a balanced dataset for training.
    \item \textbf{Preprocessing}: The digits have been size-normalised and centered in a fixed-size image, making preprocessing minimal.
\end{itemize}

\begin{figure}[h]
    \centering
    \includegraphics[width=0.7\textwidth]{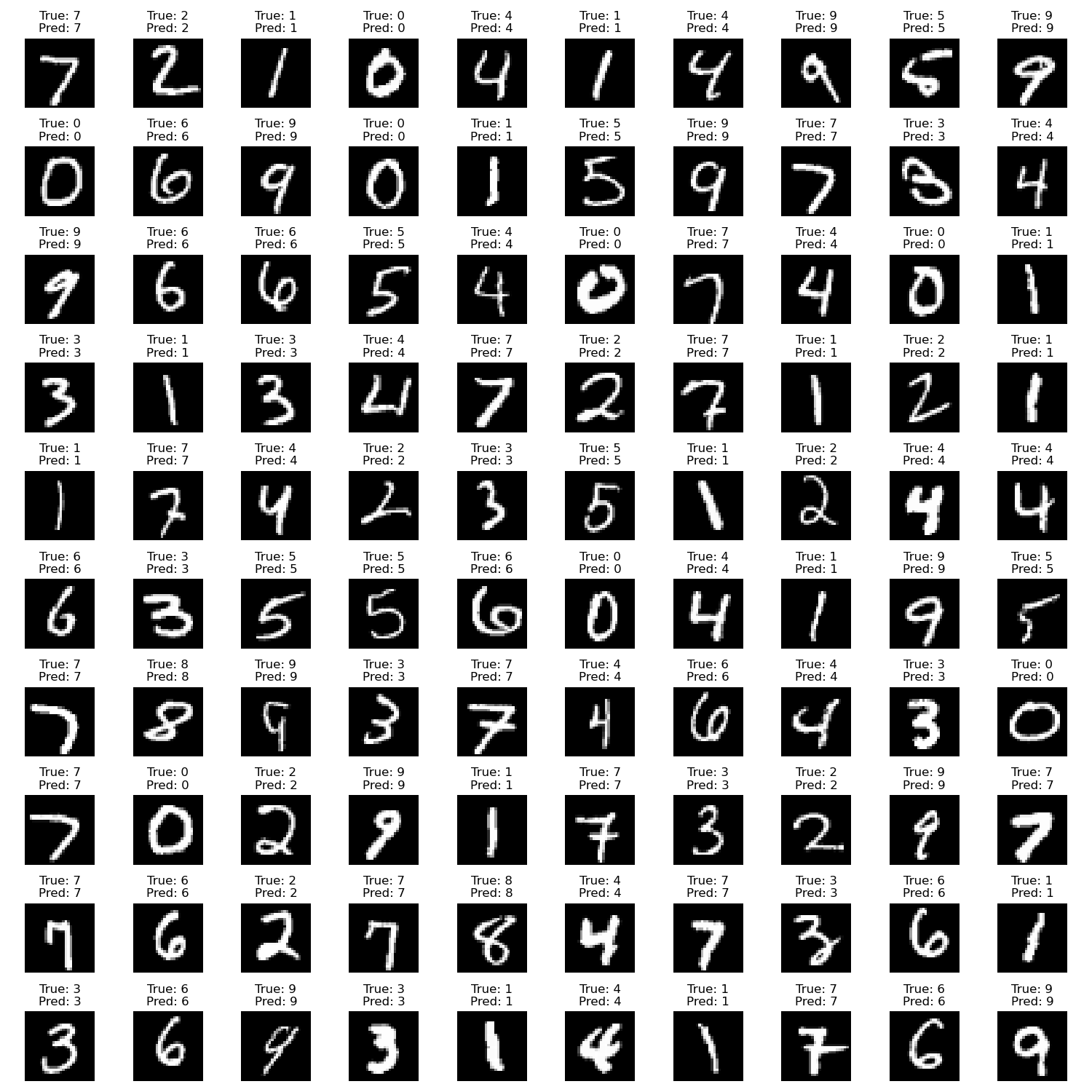}
    \caption{Visualisation of the MNIST dataset.}
    \label{fig:MNIST_fig}
\end{figure}

\Cref{fig:MNIST_fig} illustrates the MNIST dataset, showcasing examples of handwritten digits from 0 to 9. Each digit is labelled with its true class, and the predictions made by the model are displayed alongside. This figure demonstrates the digit classification task, whereby a neural network model is trained to accurately recognise and classify each digit, highlighting the role of the dataset in evaluating model performance and optimisation strategies.

The MNIST dataset provides a more complex and real-world scenario to evaluate the effectiveness of various optimisers and neural network architectures. Its simplicity, yet challenging nature, makes it a standard benchmarking dataset for new algorithms in the field of machine learning.

\begin{figure}[h]
    \centering
    \begin{tikzpicture}[
        node distance=0.7cm,
        every node/.style={draw, minimum width=2.5cm, minimum height=1cm, font=\small},
        input/.style={fill=green!20, rectangle},
        embedding/.style={fill=yellow!20, rectangle},
        attention/.style={fill=orange!20, rectangle},
        fc/.style={fill=blue!20, rectangle},
        output/.style={fill=red!20, rectangle},
        arrow/.style={-{Stealth}}
    ]

    % Nodes
    \node[input] (input) {\begin{tabular}{c} Input \\ (28x28) \end{tabular}};
    \node[embedding, right=of input] (embedding) {\begin{tabular}{c} Embedding \\ Layer (28) \end{tabular}};
    \node[attention, right=of embedding] (attention) {\begin{tabular}{c} Multi-head \\ Attention \\ (4 Heads) \end{tabular}};
    \node[fc, right=of attention] (fc1) {\begin{tabular}{c} Fully Connected \\ (128) ReLU \end{tabular}};
    \node[output, right=of fc1] (output) {\begin{tabular}{c} Output \\ (10) Log-Softmax \end{tabular}};

    % Layer Descriptions
    \node[above=0cm of embedding, align=center, draw=none] (layer1) {
        \begin{tabular}{c}
            Layer 1: \\ $\text{Linear}(28, 28)$
        \end{tabular}
    };
    \node[above=0cm of attention, align=center, draw=none] (layer2) {
        \begin{tabular}{c}
            Layer 2: \\ 
            $\text{Multi-head Attention}$ \\ 
            $(28, 28 \times 28)$
        \end{tabular}
    };
    \node[above=0cm of fc1, align=center, draw=none] (layer3) {
        \begin{tabular}{c}
            Layer 3: \\ $\text{Linear}(28 \times 28, 128)$
        \end{tabular}
    };
    \node[above=0cm of output, align=center, draw=none] (layer4) {
        \begin{tabular}{c}
            Layer 4: \\ $\text{Linear}(128, 10)$
        \end{tabular}
    };

    % Connections
    \draw[arrow] (input) -- (embedding);
    \draw[arrow] (embedding) -- (attention);
    \draw[arrow] (attention) -- (fc1);
    \draw[arrow] (fc1) -- (output);

    \end{tikzpicture}
    \caption{Neural network architecture with Multi-head Attention for the MNIST classification task}
    \label{fig:deep_neural_network}
\end{figure}
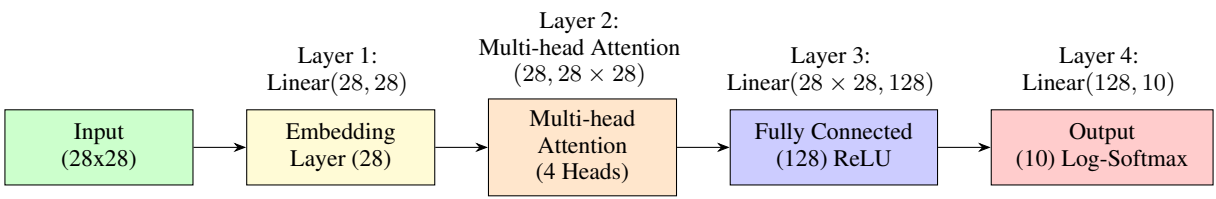

\Cref{fig:deep_neural_network} illustrates the deep neural network architecture deployed with multiple layers, including a multi-head attention mechanism to handle the complexity of the MNIST dataset.

Hyperparameters were optimised for each optimiser to ensure a fair comparison.

The performance of the newly developed optimiser, AdamZ, was rigorously evaluated against well-established optimisers such as Adam, Stochastic Gradient Descent (SGD), and RMSprop. This evaluation provided a comprehensive benchmark for assessing improvements in model classification accuracy, training time, and loss minimisation. The experiments spanned five epochs of training, each comprising of ~1,000 steps, with 100 simulations to account for randomness in parameter initialisation. These tests were conducted on a high-performance computer cluster equipped with four A100 GPUs, ensuring robust computational support for the experiments.

\begin{figure}[h]
    \centering
    \includegraphics[width=0.8\textwidth]{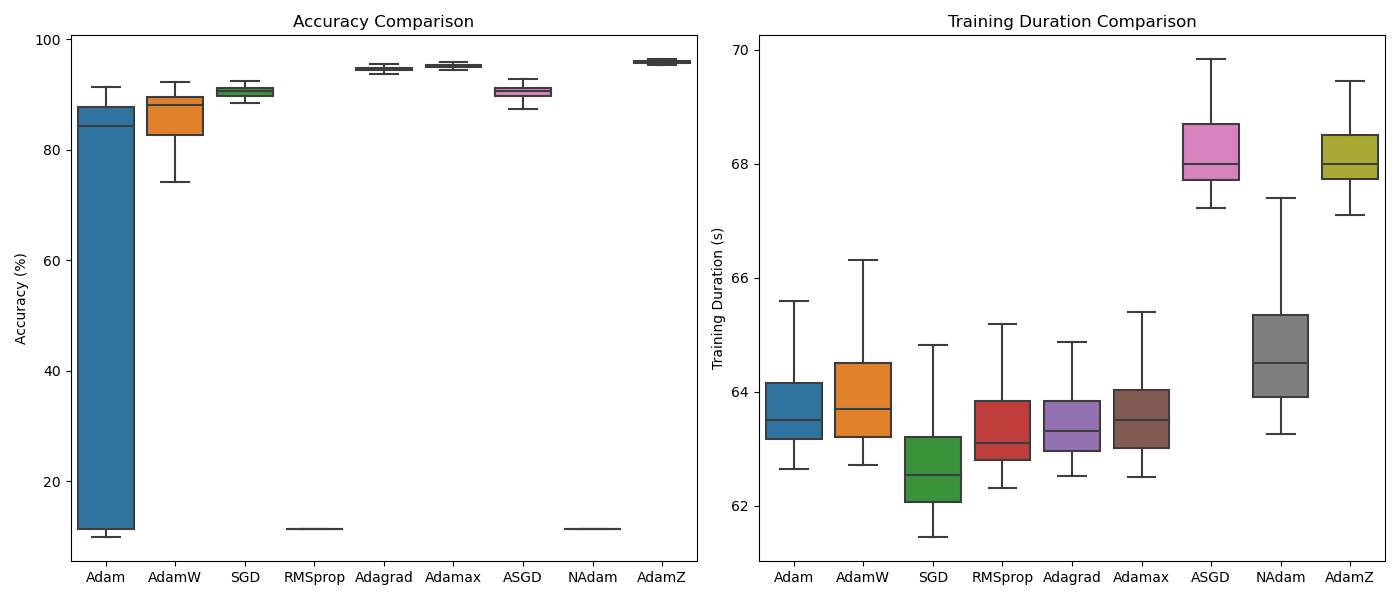}
    \caption{Accuracy and training duration on the MNIST dataset.}
    \label{fig:comparison_charts_mnist}
\end{figure}

As depicted in \Cref{fig:comparison_charts_mnist} and summarised in \Cref{tab:mnist_results}, of all the optimisers tested, AdamZ achieved the highest classification accuracy. However, it required slightly more training time, of the order of seconds, compared to the other methods. This trade-off highlights the effectiveness of AdamZ in enhancing accuracy, albeit with a marginal increase in computational time.

\begin{table}[H]
    \centering
    \begin{tabular}{|l|ccc|ccc|}
        \hline
        \textbf{Optimiser} & \multicolumn{3}{c|}{\textbf{Accuracy}} & \multicolumn{3}{c|}{\textbf{Training Duration}} \\
        \cline{2-7}
        & \textbf{Q1} & \textbf{Median} & \textbf{Q4} & \textbf{Q1} & \textbf{Median} & \textbf{Q4} \\
        \hline
        Adam & 11.35 & 84.23 & 87.73 & 63.17 & 63.50 & 64.15 \\
        AdamW & 82.58 & 88.15 & 89.50 & 63.20 & 63.69 & 64.51 \\
        SGD & 89.74 & 90.58 & 91.21 & 62.06 & 62.54 & 63.21 \\
        RMSprop & 11.35 & 11.35 & 11.35 & 62.80 & 63.11 & 63.84 \\
        Adagrad & 94.34 & 94.60 & 94.87 & 62.96 & 63.30 & 63.84 \\
        Adamax & 94.88 & 95.13 & 95.35 & 63.01 & 63.51 & 64.03 \\
        ASGD & 89.69 & 90.58 & 91.22 & 67.71 & 68.00 & 68.70 \\
        NAdam & 11.35 & 11.35 & 11.38 & 63.90 & 64.51 & 65.34 \\
        \textbf{AdamZ} & 95.74 & 95.90 & 96.03 & 67.74 & 67.99 & 68.50 \\
        \hline
    \end{tabular}
    \caption{Performance comparison of optimisers on the MNIST dataset.}
    \label{tab:mnist_results}
\end{table}

\begin{figure}[h]
    \centering
    \includegraphics[width=0.8\textwidth]{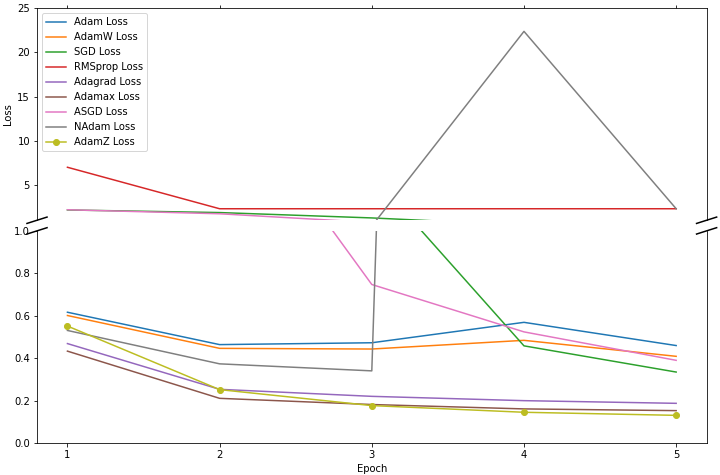}
    \caption{Optimisers' loss evolution for the MNIST dataset.}
    \label{fig:mnist_loss}
\end{figure}

Looking at the training loss evolution of the optimisers in \Cref{fig:mnist_loss}, it can be seen that AdamZ has minimised the loss faster than other optimisers.

\subsection{Analysis of Results}

The experimental findings indicate that AdamZ achieved higher model accuracy by more effectively minimising the loss, particularly on more challenging tasks such as MNIST. However, AdamZ incurred higher computational costs, indicating a trade-off that must be considered in practical applications.

Overall, these findings suggest that AdamZ is a promising candidate for applications requiring high accuracy and reliability in model predictions at the expense of a marginal increase in latency which could be tolerated in most applications. Future development of AdamZ could explore further optimisation of hyperparameters and potential enhancements in computational efficiency to reduce training time without compromising accuracy. The results from these experiments not only validate the effectiveness of AdamZ but also pave the way for its application to more diverse and challenging machine-learning tasks.

% stop

\section{Conclusions and Future Work}

\subsection{Conclusions}

The development and evaluation of the AdamZ optimiser have demonstrated its potential as a robust tool for enhancing neural network training. By dynamically adjusting the learning rate to limit overshooting and stagnation, AdamZ effectively improves convergence stability and model accuracy. The experimental results underscore the superior performance of this optimiser in minimising loss and achieving higher accuracy, particularly in complex datasets such as MNIST. Despite its slightly longer training times, the ability of AdamZ to maintain optimal learning rates positions it as a valuable asset in applications where precision is critical.

The comparative analysis with well-established optimisers, such as Adam, SGD, and RMSprop, demonstrated the AdamZstrengths in navigating the intricate landscapes of neural network training. The advanced mechanisms of the optimiser for a learning rate adjustment, guided by hyperparameters such as overshoot and stagnation factors, thresholds, and patience levels, provide a dynamically responsive but tightly controlled approach that enhances its adaptability and efficiency.

\subsection{Future Work}

Future research will focus on several key areas to further enhance the capabilities of AdamZ. Firstly, optimising the computational efficiency of AdamZ is crucial to reduce training times without compromising optimisation performance. This might require exploring alternative strategies for dynamic learning rate adjustment or integrating more advanced computational techniques.

Another promising direction is the exploration of adaptive hyperparameter tuning. Developing methods to automatically adjust hyperparameters in response to the evolving dynamics of the training process could further improve the performance and ease of use of the optimiser.

Additionally, expanding the application of AdamZ to more diverse and challenging machine learning tasks will be a priority. This includes testing its effectiveness in different neural network architectures and across various domains, such as natural language processing and computer vision, to validate its generalisability and robustness.

Finally, integrating AdamZ with emerging technologies, such as reinforcement learning frameworks or hybrid optimisation models, could open new avenues for innovation. By leveraging the strengths of AdamZ in conjunction with other optimisation strategies, it may be possible to achieve even greater improvements in model training and performance.

In conclusion, AdamZ represents a significant advancement in optimisation techniques, offering a more responsive and effective approach to learning rate adjustment. Continued research and development will ensure its relevance and utility in the ever-evolving landscape of neural network training.

\section*{Acknowledgments}
The authors gratefully acknowledge the computing time granted by the
KISSKI project.

%Bibliography
\bibliographystyle{unsrt}  
\bibliography{references}  

\end{document}